\documentclass[runningheads]{llncs}
\usepackage{bm}
\usepackage{amssymb}
\usepackage{amsfonts}
\usepackage{graphicx}
\usepackage{float}
\usepackage{lipsum}
\usepackage{multirow} % for multi columns in table
\usepackage{color, colortbl} % table colors
\definecolor{Gray}{gray}{0.95}
\usepackage{booktabs} % To thicken table lines
\newcolumntype{Y}{>{\centering\arraybackslash}X} % Center
\usepackage{adjustbox} % table full page width

\usepackage{amsmath, amsthm, amssymb, amsfonts}

\usepackage{color}

\DeclareMathOperator{\EX}{\mathbb{E}}% expected value

\usepackage{microtype}
\usepackage{graphicx}
\usepackage{subfigure}
\usepackage{booktabs} % for professional tables

\usepackage{hyperref}

\usepackage{graphicx}
\begin{document}
\title{Deep Conditional Transformation Models}%\thanks{Supported by organization x.}}
\author{
Philipp F.M. Baumann\inst{1}\orcidID{0000-0001-8066-1615} 
\and
Torsten Hothorn\inst{2}\orcidID{0000-0001-8301-0471}
\and
David Rügamer\inst{3}\orcidID{0000-0002-8772-9202}
}

\authorrunning{Baumann et al.}
\institute{KOF Swiss Economic Institute, ETH Zurich, Zurich, Switzerland\\
\email{baumann@kof.ethz.ch}
\and
Epidemiology, Biostatistics and Prevention Institute, University of Zurich, Zurich, Switzerland\\
\email{torsten.hothorn@uzh.ch}\\
\and
Department of Statistics, LMU Munich, Munich, Germany\\
\email{david.ruegamer@stat.uni-muenchen.de}}
\maketitle              % typeset the header of the contribution
\begin{abstract}

Learning the cumulative distribution function (CDF) of an outcome variable conditional on a set of features remains challenging, especially in high-dimensional settings. Conditional transformation models provide a semi-parametric approach that allows to model a large class of conditional CDFs without an explicit parametric distribution assumption and with only a few parameters. Existing estimation approaches within this class are, however, either limited in their complexity and applicability to unstructured data sources such as images or text, lack interpretability, or are restricted to certain types of outcomes. We close this gap by introducing the class of deep conditional transformation models which unifies existing approaches and allows to learn both interpretable (non-)linear model terms and more complex neural network predictors in one holistic framework. To this end we propose a novel network architecture, provide details on different model definitions and derive suitable constraints as well as network regularization terms. We demonstrate the efficacy of our approach through numerical experiments and applications. %Its efficacy and competitiveness is demonstrated through numerical experiments and an application to the movie reviews data set from Kaggle.  

\end{abstract}
%Transformation Models
%Structured Effects
%Distributional Regression
%Normalizing Flows
%Constrained Estimation
%
%

\section{Introduction}

%[To be written: Uncertainty Quantification $\rightarrow$ Aleatoric uncertainty $\rightarrow$ Distributional Regression $\rightarrow$ Non-Parametric vs Semi-Parametric Estimation $\rightarrow$ Trafo Models] [Cite ICML, AISTATS Papers that deal with dist. learning and Aleatoric uncertainty]

%Many machine learning applications call for uncertainty quantification when predicting the most probable class label conditional on a set of features or when finding an estimate for the conditional expected value of a continuous response. While tools from statistical and machine learning have shown astonishing performance regarding these tasks since the 2000s \cite[e.g.][]{Geurts.2006, Hothorn.2006, Huang.2006}, learning the full predictive distribution to incorporate this uncertainty has only gained attention in the more recent years \cite{Chernozhukov.2013, Gneiting.2014, Cabrera.2017, Hothorn.2020}. The idea is not new, since, for example, the full predictive distribution can even be obtained from a (classical) linear regression model by introducing additional parametric assumptions as for the error variance. However, if we do not want to impose such parametric restrictions for the conditional distributions but let the model coefficients vary with the response conditional on the features to determine the shape of the distribution we arrive at distribution regression as introduced by \cite{Foresi.1995}. The most recent literature \cite{Hothorn.2020} distinguishes between four approaches to estimate the conditional distribution. 

Recent discussions on the quantification of uncertainty have emphasized that a distinction between aleatoric and epistemic uncertainty is useful in classical machine learning \cite{Senge.2014,Hullermeier.2019}. Moreover, this distinction was also advocated in the deep learning literature \cite{Kendall.2017,Depeweg.2018}.While epistemic uncertainty describes the uncertainty of the model and can be accounted for in a Bayesian neural network, aleatoric uncertainty \cite{Horal.1996} can be captured by modeling an outcome probability distribution that has a stochastic dependence on features (i.e., conditional on features). Apart from non-parametric estimation procedures, four fundamental approaches in statistics exist that allow to model the stochastic dependence between features and the outcome distribution \cite{Hothorn.2020}. First, parametric models where additive functions of the features describe the location, scale and shape (LSS) of the distribution \cite{Rigby.2005} or where these features are used in heteroscedastic Bayesian additive regression tree ensembles \cite{Pratola.2017}. Second, quantile regression models \cite{Koenker.2005,Meinshausen.2006,Athey.2019} that directly model the conditional quantiles with a linear or non-linear dependence on feature values. Third, distribution regression and transformation models \cite{Foresi.1995,Rothe.2013,Chernozhukov.2013,Wu.2013,Leorato.2015} that have response-varying effects on the probit, logit or complementary log-log scale. Finally, hazard regression \cite{Kooperberg.1995} which estimates a non-proportional hazard function conditional on feature values. Parallel to this, various approaches in machine learning and deep learning have been evolved to model the outcome distribution through input features. %While some deep learning approaches combine statistical concepts and neural networks \cite[e.g.,][ embedding LSS models in a neural network]{Ruegamer.2020}, 
%These parallel streams of research sometimes independently follow a very similar idea. 
A prominent example is normalizing flows (see, e.g. \cite{Tabak.2013,Rezende.2015,Papamakarios.2019,Kobyzev.2020}), used to learn a complex distribution of an outcome based on feature values. Normalizing flows start with a simple base distribution $F_Z$ and transform $F_Z$ to a more complex target distribution using a bijective transformation of the random variable coming from the base distribution. The most recent advances utilize monotonic polynomials \cite{Priyank.2019,Ramasinghe.2021} or splines \cite{muller.2019,Durkan.2019,Durkan.2019.b} to learn this transformation. As pointed out recently by several authors \cite{Klein.2019,Sick.2020}, normalizing flows are conceptually very similar to transformation models. However, normalizing flows, in contrast to transformation models, usually combine a series of transformations with limited interpretability of the influence of single features on the distribution of the target variable. In this work, we instead focus on conditional transformation approaches that potentially yield as much flexibility as normalizing flows, but instead of defining a generative approach, strive to build interpretable regression models without too restrictive parametric distribution assumptions.  % such that this literature can to some extent be attributed to the third approach mentioned by \cite{Hothorn.2020}.
% Other close connections between statistical models and deep learning have been established allowing the estimation of a feature dependent outcome distribution, e.g., by embedding LSS approaches into a neural network \cite{Ruegamer.2020}. %The advancements of \cite{Sick.2020} did not regard the interpretability of feature effects either. 
%We will now give an introduction into the class of transformation models before we specify our contribution in more detail.
% These approaches can bypass restrictions in the analysis with regression models in complex or high-dimensional data settings or restrictions due to software limitations. Extensions to regression models with additional deep learning predictor as in \cite{Ruegamer.2020} is another advantage we want to pursue in this work. 

% \cite{Sick.2020} also extend the class of transformation models to allow for deep neural networks as driving input of the transformation. 

% Conditional expectations and quantiles are readily derived from transformation models on the outcome's original scale.
%Classical machine learning (e.g. random forests) are concerned with changes in the mean rather than in higher moments.

\subsection{Transformation Models}

The origin of transformation models (TM) can be traced back to \cite{Box.1964} studying a parametric approach to transform the variable of interest $Y$ prior to the model estimation in order to meet a certain distribution assumption of the model.
Many prominent statistical models, such as the Cox proportional hazards model or the proportional odds model for ordered outcomes, can be understood as transformation models. Estimating transformation models using a neural network has been proposed by \cite{Sick.2020}. However, \cite{Sick.2020} only focus on a smaller subclass of transformation models, we call (linear) shift transformation models and on models that are not interpretable in nature. 
Recently, fully parameterized transformation models have been proposed  \cite{Hothorn.2014,mlt.2018} which employ likelihood-based learning to estimate the cumulative distribution function $F_Y$ of $Y$ via estimation of the corresponding transformation of $Y$. The main assumption of TM is that $Y$ follows a known, log-concave error distribution $F_Z$ after some monotonic transformation $h$. CTMs specify this transformation function conditional on a set of features $\bm{x}$:
\begin{equation} \label{eq:CTM}
\mathbb{P}(Y \leq y | \bm{x}) = F_{Y|\bm{x}}(y) = F_Z(h(y | \bm{x})).    
\end{equation}
%Linear transformation models additionally assume that this transformed response is linearly associated with the set of features.
The transformation function $h$ can be decomposed as $h(y | \bm{x}) := h_1 + h_2$, where $h_1$ and $h_2$ can have different data dependencies as explained in the following. When $h_1$ depends on $y$ as well as $\bm{x}$, we call the CTM an \emph{interacting CTM}. When $h_1$ depends on $y$ only, we call the model a \emph{shift CTM}, with shift term $h_2$. When $h_2$ is omitted in an interacting CTM, we call the CTM a \emph{distributional CTM}. In general, the bijective function $h(y | \bm{x})$ is unknown a priori and needs to be learned from the data.
\cite{mlt.2018} study the likelihood of this transformation function and propose an estimator for the most likely transformation. \cite{mlt.2018} specify the transformation function through a flexible basis function approach, which, in the unconditional case $h(y)$ (without feature dependency), is given by $h(y) = \bm{a}(y)^\top \bm{\vartheta}$ where $\bm{a}(y)$ is a matrix of evaluated basis functions and $\bm{\vartheta}$ a vector of basis coefficients which can be estimated by maximum likelihood. For continuous $Y$ Bernstein polynomials \cite{Farouki.2012} with higher order $M$ provide a more  flexible but still computationally attractive choice for $\bm{a}$. That is,
\begin{equation} \label{eq:BSP}
   \bm{a}(y)^\top \bm{\vartheta} = \frac{1}{(M + 1)}\sum_{m = 0}^{M} \vartheta_m f_{Be(m + 1, M - m + 1)}(\tilde{y})
\end{equation}
where $f_{Be(m,M)}$ is the probability density function of a Beta distribution with parameters m, M and a normalized outcome $\tilde{y} := \frac{y - l}{u-l} \in [0,1]$ with $u > l$ and $u,l \in \mathbb{R}$. In order to guarantee monotonicity of the estimate of $F_{Y|\bm{x}}$, strict monotonicity of $\bm{a}(y)^\top \bm{\vartheta}$ is required. This can be achieved by restricting $\vartheta_m > \vartheta_{m-1}$ for $m = 1, \ldots,  M + 1$. 
%h monotonicity is not a assumptions but a prerequisite to gain a fully stochastic model. 
When choosing $M$, the interplay with $F_Z$ should be considered. For example, when $F_Z = \Phi$, the standard Gaussian distribution function, and $M = 1$, then $\hat{F}_{Y}$ will also belong to the family of Gaussian distributions functions. Further, when we choose $M = n - 1$ with $n$ being the number of independent observations, then $\hat{F}_{Y}$ is the non-parametric maximum likelihood estimator which converges to $F_{Y}$ by the Glivenko-Cantelli lemma \cite{HothornJSS.2020}. As a result, for small $M$ the choice of $F_Z$ will be decisive, while TMs will approximate the empirical cumulative distribution function well when $M$ is large independent of the choice of $F_Z$.  Different choices for $F_Z$ have been considered in the literature (see, e.g., \cite{mlt.2018}), such as the standard Gaussian distribution function ($\Phi$), the standard logistic distribution function ($F_{L}$) or the minimum extreme value distribution function ($F_{MEV}$).

In CTMs with structured additive predictors (STAP), features considered in $h_1$ and $h_2$ enter through various functional forms and are combined as an additive sum. The STAP is given by 
\begin{equation} \label{eq:STAP}
   \eta_{struc} = s_1(\bm{x}) + \ldots + s_k(\bm{x})
\end{equation}
with $s_1, \ldots, s_k$ being partial effects of one or more features in $\bm{x}$. Common choices include linear effects $\bm{x}^\top \bm{w}$ with regression coefficient $\bm{w}$ and non-linear effects based on spline basis representation, spatial effects, varying coefficients, linear and non-linear interaction effects or individual-specific random effects \cite{Fahrmeier.2013}. Structured additive models have been proposed in many forms, for example in additive (mixed) models where $\EX(Y|\bm{x}) = \eta_{struc}$.

\subsection{Related Work and Our Contribution}\label{sec:contribution}

The most recent advances in transformation models \cite{Hothorn.2017,Klein.2019,Hothorn.2020} %can not learn the CDF and thus potentially complex $h_1$ and $h_2$ simultaneously without explicit prior specification of the features while still retaining interpretability. These algorithms 
learn the transformation functions $h_1$ an $h_2$ separately, using, e.g., a model-based boosting algorithm with pre-specified base learners \cite{Hothorn.2020}. Very recent neural network-based approaches allow for the joint estimation of both transformation functions, but do either not yield interpretable models \cite{Sick.2020} or are restricted to STAP with ordinal outcomes \cite{Kook.2020}. %or do not incorporate a STAP in $h_2$ \cite{HothornJSS.2020}.

Our framework combines the existing frameworks and thereby extends approaches for continuous outcomes to transformation models able to 1) learn more flexible and complex specifications of $h_1$ and $h_2$ simultaneously 2) learn the CDF without the necessity of specifying the (type of) feature contribution a priori, 3) retain the interpretability of the structured additive predictor in $h_1$ and $h_2$ 4) estimate structured effects in high-dimensional settings due to the specification of the model class within a neural network 5) incorporate unstructured data source such as texts or images.

\section{Model and Network Definition}

We now formally introduce the deep conditional transformation model (DCTM), explain its network architecture and provide details about different model definitions, penalization and model tuning.

\subsection{Model Definition}\label{sec:modDef}
\begin{figure*}[htb]
    \centering
    \includegraphics[page=7,
    trim=1cm 1cm 1cm 0.5cm, 
    width = 0.83\textwidth]{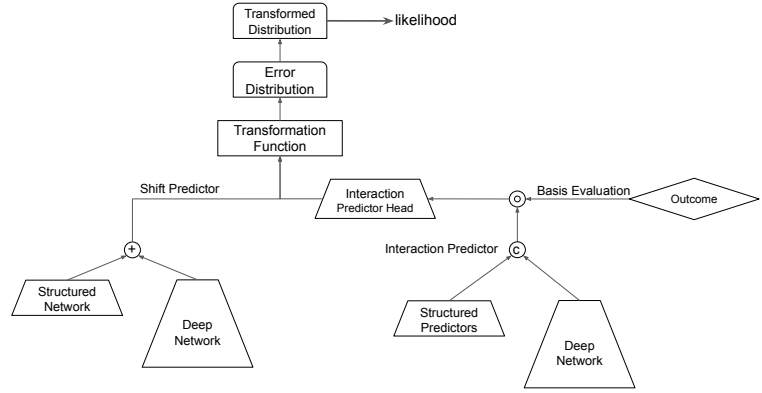}
    \caption{Architecture of a deep conditional transformation model. Both the shift and interaction predictor can potentially be defined by a structured network including linear terms, (penalized) splines or other structured additive regression terms and deep neural network defined by an arbitrary network structure. While the shift predictor ($\bm{\mathcal{C}}\bm{\Psi}$) is a sum of both subnetwork predictions, the interaction predictor ($\bm{\mathcal{A}}\odot\bm{\mathcal{B}}$) is only multiplied with a final 1-hidden unit fully-connected layer (network head, $\text{vec}(\bm{\Gamma})$) after the structured predictors and latent features of the deep neural network are combined with the basis evaluated outcome. The shift and interaction network part together define the transformation function, which transforms the error distribution and yields the final likelihood used as loss function.}
    \label{fig:arch}
    \vskip -0.1in
\end{figure*}
Following \cite{Hothorn.2020}, we do not make any explicit parameterized distribution assumption about $Y$, but instead assume 
\begin{equation} \label{eq:modelassumption}
\mathbb{P}(Y \leq y | \bm{x}) = F_Z(h(y | \bm{x}))    
\end{equation}
with error distribution $F_Z : \mathbb{R} \mapsto [0,1]$, an a priori known CDF that represents the data generating process of the transformed outcome $h(Y | \bm{x})$ conditional on some features $\bm{x} \in \chi$. For tabular data, we assume $\bm{x} \in \mathbb{R}^p$. For unstructured data sources such as images, $\bm{x}$ may also include  multidimensional inputs. Let $f_Z$ further be the corresponding probability density function of $F_Z$. We model this transformation function conditional on some predictors $\bm{x}$ by $h(y | \bm{x}) = h_1 + h_2 = \bm{a}(y)^\top \bm{\vartheta}(\bm{x}) + \beta(\bm{x})$, where $\bm{a}(y)$ is a (pre-defined) basis function $\bm{a} : \Xi \mapsto \mathbb{R}^{M+1}$ with $\Xi$ the sample space and $\bm{\vartheta} : \chi_\vartheta \mapsto \mathbb{R}^{M+1}$ a conditional parameter function defined on $\chi_\vartheta \subseteq \chi$. $\bm{\vartheta}$ is parameterized through structured predictors such as splines, unstructured predictors such as a deep neural network, or the combination of both and $\beta(\bm{x})$ is a feature dependent distribution shift. More specifically, we model $\bm{\vartheta}(\bm{x})$ by the following additive predictor:
\begin{equation} \label{eq:vartheta}
 \bm{\vartheta}(\bm{x}) =  \sum_{j=1}^J \bm{\Gamma}_j \bm{b}_j(\bm{x}),  
\end{equation}
with $\bm{\Gamma}_j \in \mathbb{R}^{(M+1) \times O_j}, O_j \geq 1,$ being joint coefficient matrices for the basis functions in $\bm{a}$ and the chosen predictor terms $\bm{b}_j : \chi_{b_j} \mapsto \mathbb{R}^{O_j}, \chi_{b_j} \subseteq \chi$. We allow for various predictor terms including an intercept (or bias term), linear effects $\bm{b}_j(\bm{x}) = x_{k_j}$ for some $k_j \in \{1, \ldots, p\}$, structured non-linear terms $\bm{b}_j(\bm{x}) = G(x_{k_j})$ with some basis function $G : \mathbb{R} \mapsto \mathbb{R}^q, q \geq 1$ such as a B-spline basis, bivariate non-linear terms $\bm{b}_j(\bm{x}) = G'(x_{k_j},x_{k_{j'}})$ using a tensor-product basis $G': \mathbb{R} \times \mathbb{R} \mapsto \mathbb{R}^{q'}, q' \geq 1$ or neural network predictors $\bm{b}_j(\bm{x}) = d(\bm{x}_{k_{j}})$, which define an arbitrary (deep) neural network that takes (potentially multidimensional) features $\bm{x}_{k_j} \in \chi$. The network will be used to learn latent features representing the unstructured data source. These features are then combined as a linear combination when multiplied with $\bm{\Gamma}_j$. The same types of predictors can also be defined for the shift term $\beta(\bm{x}) = \sum_{j=1}^J \bm{c}_j(\bm{x})^\top \bm{\psi}_j$, which we also defined as an additive predictor of features, basis functions or deep neural networks times their (final) weighting $\bm{\psi}_j$.

The final model output for the transformation of $y$ is then given by 
\begin{equation} \label{eq:modeloutput}
\bm{a}(y)^\top \bm{\vartheta}(\bm{x}) =  \bm{a}(y)^\top \bm{\Gamma} \bm{B},  
\end{equation}
with $\bm{\Gamma} = (\bm{\Gamma}_1,\ldots,\bm{\Gamma}_J) \in \mathbb{R}^{(M+1) \times P}, P = \sum_{j=1}^J O_j$ the stacked coefficient matrix combining all $\bm{\Gamma}_j$s and $\bm{B} \in \mathbb{R}^{P}$ a stacked vector of the predictor terms $b_j(\bm{x})$s. Based on model assumption \eqref{eq:modelassumption} we can define the loss function based on the change of variable theorem 
$$
    f_Y(y | \bm{x}) = f_Z(h(y | \bm{x})) \cdot \left| \frac{\partial h(y|\bm{x})}{\partial y}\right|
$$
as 
\begin{equation} \label{eq:loss}
\begin{split}
 \ell&(h(y|\bm{x})) = - \log f_Y(y | \bm{\vartheta}(\bm{x}), \bm{\beta}(\bm{x}))\\
 &= -\log f_Z (\bm{a}(y)^\top \bm{\vartheta}(\bm{x}) + \beta(\bm{x})) - \log[\bm{a}'(y)^\top \bm{\vartheta}(\bm{x})]
\end{split}
\end{equation}
with $\bm{a}'(y) = \partial \bm{a}(y) / \partial y$. 

\begin{figure}
\begin{center}
    \includegraphics[width = 0.8\columnwidth, height=0.15\textheight]{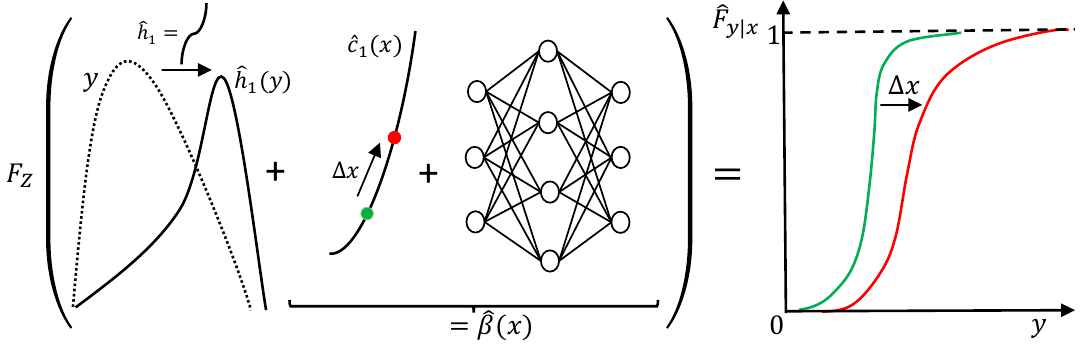}
    \caption{Schematic representation of an exemplary DCTM with a learned transformation $\hat{h}_1$ for the outcome $y$. %which does not interact with $x$ here.
    The shift term $\hat{\beta}(x)$ is composed of an estimated smooth term $\hat{c}_1(x) = c_1(x)\hat{\psi}_1$ for $x$ and a neural network predictor. An increase in $x$ is indicated by $\Delta x$ with corresponding effect on $\hat{F}_{Y|x}$ through $\hat{h}_2 = \hat{\beta}(x)$ on the right hand side of the equation.}
    \label{fig:DCT}    
\end{center}
    \vskip -0.1in
\end{figure}

For $n$ observations $(y_i,\bm{x}_i), i = 1,\ldots,n$, we can represent \eqref{eq:modeloutput} as 
\begin{equation} \label{eq:rowwisetp}
(\bm{\mathcal{A}} \odot \bm{\mathcal{B}}) \text{vec}(\bm{\Gamma}^\top)
\end{equation}
with $\bm{\mathcal{A}} = (\bm{a}(y_1), \ldots, \bm{a}(y_n))^\top \in \mathbb{R}^{n \times (M+1)}$, $\bm{\mathcal{B}} = (\bm{B}_1,\ldots,\bm{B}_n)^\top \in \mathbb{R}^{n \times P}$, vectorization operator $\text{vec}(\cdot)$ and the row-wise tensor product (also known as transpose Kathri-Rao product) operator $\odot$. Similar, the distribution shift can be written in matrix form as $\bm{\mathcal{C}}\bm{\Psi}$ with $\bm{\mathcal{C}} \in \mathbb{R}^{n \times Q}$ consisting of the stacked $\bm{c}_j(\bm{x})$s and $\bm{\Psi} = (\bm{\psi}_1^\top, \ldots, \bm{\psi}_J^\top)^\top \in \mathbb{R}^Q$ the stacked vector of all shift term coefficients. A schematic  representation of an exemplary DCTM is given in Figure~\ref{fig:DCT}.

\subsection{Network Definition}

Our network consists of two main parts: a feature transforming network (FTN) part, converting $\bm{X} = (\bm{x}_1^\top,\ldots,\bm{x}_n^\top)^\top \in \mathbb{R}^{n \times p}$ to
$\bm{\mathcal{B}}$ and an outcome transforming network (OTN) part, transforming $\bm{y} = (y_1,\ldots,y_n)^\top \in \mathbb{R}^n$ to $h(\bm{y} | \bm{X}) \in \mathbb{R}^n$. In the OTN part the matrix $\bm{\Gamma}$ is learned, while the FTN part only contains additional parameters to be learned by the network if some feature(s) are defined using a deep neural network. In other words, if only structured linear effects or basis function transformations are used in the FTN part, $\bm{\Gamma}$ contains all trainable parameters. Figure~\ref{fig:arch} visualizes an exemplary architecture.

After the features are processed in the FTN part, the final transformed outcome is modeled using a conventional fully-connected layer with input $\bm{\mathcal{A}} \odot \bm{\mathcal{B}}$, one hidden unit with linear activation function and weights corresponding to $\text{vec}( \bm{\Gamma} )$. The deep conditional transformation model as visualized in Figure~\ref{fig:arch} can also be defined with one common network which is split into one part that is added to the shift predictor and one part that is used in the interaction predictor.   %Using the definition \eqref{eq:modeloutput} for model estimation \cite[also known as linear array framework,][]{Currie.2006}, however, is more efficient in terms of number of multiplication operations and does not require to construct the row-wise tensor product explicitly \cite[cf.][]{Hothorn.2020}. It can be implemented using $(\bm{\mathcal{A}} \ast (\bm{\mathcal{B}} \bm{\Gamma}^\top)) \bm{1}_{M+1} $ with Hadamard product $\ast$.

\subsection{Penalization} \label{sec:penalty}
$L_1$-, $L_2$-penalties can be incorporated in both the FTN and OTN part by adding corresponding penalty terms to the loss function. We further use smoothing penalties for structured non-linear terms by regularizing the respective entries in $\bm{\Psi}$ and $\bm{\Gamma}$ to avoid overfitting and easier interpretation. Having two smoothing directions, the penalty for $\bm{\Gamma}$ is constructed using a Kronecker sum of individual marginal penalties for anisotropic smoothing $$\bm{D}_{\Gamma} = \lambda_a \bm{D}_a \oplus \lambda_b \bm{D}_b,$$ where the involved tuning parameters $\lambda_a, \lambda_b$ and penalty matrices $\bm{D}_a, \bm{D}_b$ correspond to the direction of $y$ and the features $\bm{x}$, respectively. Note, however, that for $\bm{\Gamma}$, the direction of $y$ usually does not require additional smoothing as it is already regularized through the monotonicity constraint \cite{mlt.2018}. The corresponding penalty therefore reduces to 
\begin{equation} \label{eq:pen}
\bm{D}_{\Gamma} = \bm{I}_P  \otimes (\lambda_b \bm{D}_b)   
\end{equation}
with the diagonal matrix $\bm{I}_P$ of size $P$. These penalties are added to the negative log-likelihood defined by \eqref{eq:loss}, e.g., $$\ell_{pen} = \ell(h(y|\bm{x})) + \text{vec}(\bm{\Gamma})^\top \bm{D}_{\Gamma} \text{vec}(\bm{\Gamma})$$ for a model with penalized structured effects only in $\bm{\mathcal{B}}$. As done in \cite{Ruegamer.2020} we use the Demmler-Reinsch orthogonalization to relate each tuning parameter for smoothing penalties to its respective degrees-of-freedom, which allows a more intuitive setting of parameters and, in particular, allows to define equal amount of penalization for different smooth terms. Leaving the least flexible smoothing term unpenalized and adjusting all others to have the same amount of flexibility works well in practice.

\subsection{Bijectivitiy and Monotonocity Constraints}

To ensure bijectivity of the transformation of each $y_i$, we use Bernstein polynomials for $\mathcal{A}$ and constraint the coefficients in $\bm{\Gamma}$ to be monotonically increasing in each column. The monotonicity of the coefficients in $\bm{\Gamma}$ can be implemented in several ways, e.g., using the approach by \cite{Gupta.2016} or \cite{Sick.2020} on a column-basis. Note that this constraint directly yields monotonically increasing transformation functions if $P=1$, i.e., if no or only one feature is used for $h_1$. If $P>1$, we can ensure monotonicity of $h_1$ by using predictor terms in $\bm{\mathcal{B}}$ that are non-negative. A corresponding proof can be found in the Supplement (Lemma 1). Intuitively the restriction can be seen as an implicit positivity assumption on the learned standard deviation of the error distribution $F_Z$ as described in the next section using the example of a normal distribution. Although non-negativity of predictor terms is not very restrictive, e.g., allowing for positive linear features, basis functions with positive domain such as B-splines or deep networks with positivity in the learned latent features (e.g., based on a ReLU activation function), the restriction can be lifted completely by simply adding a positive constant to $\bm{\mathcal{B}}$.

\subsection{Interpretability and Identifiability Constraints}

Several choices for $M$ and $F_Z$ will allow for particular interpretation of the coefficients learned in $\bm{\Psi}$ and $\bm{\Gamma}$. 
When choosing $F_Z = \Phi$ and $M = 1$, the DCTM effectively learns an additive regression model with Gaussian error distribution, i.e., $Y|\bm{x} \sim N(\tilde{\beta}(\bm{x}), \sigma_s^2)$. The unstandardized structured additive effects in $\tilde{\beta}(\bm{x})$ can then be divided by $\sigma_s$ yielding $\beta(\bm{x})$. Therefore $\beta(\bm{x})$ can be interpreted as shifting effects of normalized features on the transformed response $\mathbb{E}(h_1(y)|\bm{x})$. 
For $M > 1$, features in $\beta(\bm{x})$ will also affect higher moments of $Y|\bm{x}$ through a non-linear $h_1$, leading to a far more flexible modeling of $F_{Y|\bm{x}}$. Smooth monotonously increasing estimates for $\beta(\bm{x})$ then allow to infer that a rising $\bm{x}$ leads to rising moments of $Y|\bm{x}$ %%in a two-fold non-linear way (i.e. through $h_1$ and $\bm{c}_j(\bm{x})$) 
independent of the choice for $F_Z$. Choosing $F_Z = F_{MEV}$ or $F_Z = F_{L}$ allows $\beta(\bm{x})$ to be interpreted as additive changes on the log-hazard ratio or on the log-odds ratio, respectively. %, independent of the choice of $M$.
%When encoding structured terms in $\bm{\mathcal{B}}$, 
The weights in $\bm{\Gamma}$ determine the effect of $\bm{x}$ on $F_{Y|\bm{x}}$ as well as whether $F_{Y|\bm{x}}$ varies with the values of $y$ yielding a response-varying distribution \cite{Chernozhukov.2013} or not. %The latter would suggest to encode a feature in $\bm{\mathcal{C}}$ rather than in $\bm{\mathcal{B}}$.
In general, structured effects in $\bm{\Gamma}$ are coefficients of the tensor product $\bm{\mathcal{A}} \odot \bm{\mathcal{B}}$ and can, e.g., be interpreted by 2-dimensional contour or surface plots (see, e.g., Figure~\ref{fig:struct2}).

In order to ensure identifiability and thus interpretability of structured effects in $h_1$ and $h_2$, several model definitions require the additional specifications of constraints. If certain features in $\bm{\mathcal{B}}$ or $\bm{\mathcal{C}}$ are modeled by both a flexible neural network predictor $d(\bm{x})$ and structured effects $s(\bm{x})$, the subnetwork $d(\bm{x})$ can easily assimilate effects $s(\bm{x})$ is supposed to model. In this case, identifiability can be ensured by an orthogonalization cell \cite{Ruegamer.2020}, projecting the learned effects of $d(\bm{x})$ in the orthogonal complement of the space spanned by features modeled in $s(\bm{x})$.
%In short, the orthogonalization cell separates the feature space into a subspace spanned by columns of structured terms in $\bm{\mathcal{B}}$ and $\bm{\mathcal{C}}$ while the deep neural network predictor is projected onto its complement. 
Further, when more than one smooth effect or deep neural network is incorporated in either $\bm{\mathcal{B}}$ or $\bm{\mathcal{C}}$, these terms can only be learned up to an additive constants. To solve this identifiability issue we re-parameterize the terms and learn these effects with a sum-to-zero constraint. % (i.e. $\sum_{i}\hat{e}(x_i) = 0$). 
As a result, corresponding effects can only be interpreted on a relative scale. Note that this is a limitation of additive models per se, not our framework.

\section{Numerical Experiments}

We now demonstrate the efficacy of our proposed framework for the case of a shift CTM, a distributional CTM and an interacting CTM based on a general data generating process (DGP). 

\paragraph{Data Generating Process}
The data for the numerical experiments were generated according to $g(y) = \eta(\bm{x}) + \epsilon(\bm{x})$ where $g: \mathbb{R}^n \mapsto \mathbb{R}^n$ is bijective and differentiable, $\eta(\bm{x})$ is specified as in \eqref{eq:STAP} and $\epsilon \sim F_Z$ with $F_Z$ being the error distribution. We choose $\epsilon(\bm{x}) \sim N(0,\sigma^2(\bm{x})
)$ where $\sigma^2(\bm{x}) \in \mathbb{R}^+$ is specified as in \eqref{eq:STAP} so that we can rewrite the model as
\begin{equation}\label{eq:sim}
   F_Z\left(\frac{g(y)-\eta(\bm{x})}{\sigma(\bm{x})}\right) = F_Z\left(h_1 + h_2\right).
\end{equation}
From \eqref{eq:CTM} and our model definition, \eqref{eq:sim} can be derived by defining $h_1$ as $g(y)\sigma^{-1}(\bm{x})$ and $h_2$ as $-\eta(\bm{x})\sigma^{-1}(\bm{x})$. We finally generate $y$ according to $g^{-1}(\eta(\bm{x}) + \epsilon(\bm{x}))$ with $\epsilon(\bm{x}) \sim N(0,\sigma^2(\bm{x}))$. We consider different specification for $g$, $\eta$, $\sigma$ and the order of the Bernstein polynomial $M$ for different samples sizes $n$ (see Appendix).

\paragraph{Evaluation}
To assess the estimation performance, we compute the relative integrated mean squared error (RIMSE) between $\hat{h}_1$, evaluated on a fine grid of ($y_i$, $\sigma(\bm{x}_{i})$) value pairs, with the true functional form of $h_1$ as defined by the data generating process. For the estimation performance of $h_2$, we evaluate the corresponding additive predictor by calculating the mean squared error (MSE) between estimated and true linear coefficients for linear feature effects and the RIMSE between estimated and true smooth non-linear functions for non-linear functional effects. We compare the estimation against %most likely transformation implemented in \texttt{mlt} \cite{HothornJSS.2020} and
transformation boosting machines (TBM) \cite{Hothorn.2020} %implemented in \texttt{tbm}  %both for the statistical software R \cite{R.2020}
%to compare the estimation performance against existing methods
that also allow to specify structured additive predictors. Note, however, that TBMs only implement either the shift (TBM-Shift) or distributional CTM (TBM-Distribution), but do not allow for the specification of an interacting CTM with structured predictors, a novelty of our approach. In particular, only the TBM-Shift comes with an appropriate model specification such that it can be used for comparison in some of the DGP defined here.

\paragraph{Results}

We first discuss the 4 out of 10 specifications of the true DGP where $h_1$ is not learned through features and thus allows for a direct comparison of TBM-Shift and DCTMs. For $h_1$, we find that, independent of the size of the data set and the order of the Bernstein polynomial, DCTMs provide a viable alternative to TBM-Shift, given the overlap between the (RI)MSE distributions and the fact that the structured effects in DCTMs are not tuned extensively in these comparisons. % are on par with  TBM-Shift when taking the distribution across all replications into account. 
For $h_2$, DCTMs outperform TBM-Shift in all 16
configurations for $M$/$n$ among the 4 DGPs depicted in Figure~\ref{fig:struct} when taking the mean or the median across the 20 replications.
\begin{figure}
    \centering
    \includegraphics[width = \columnwidth]{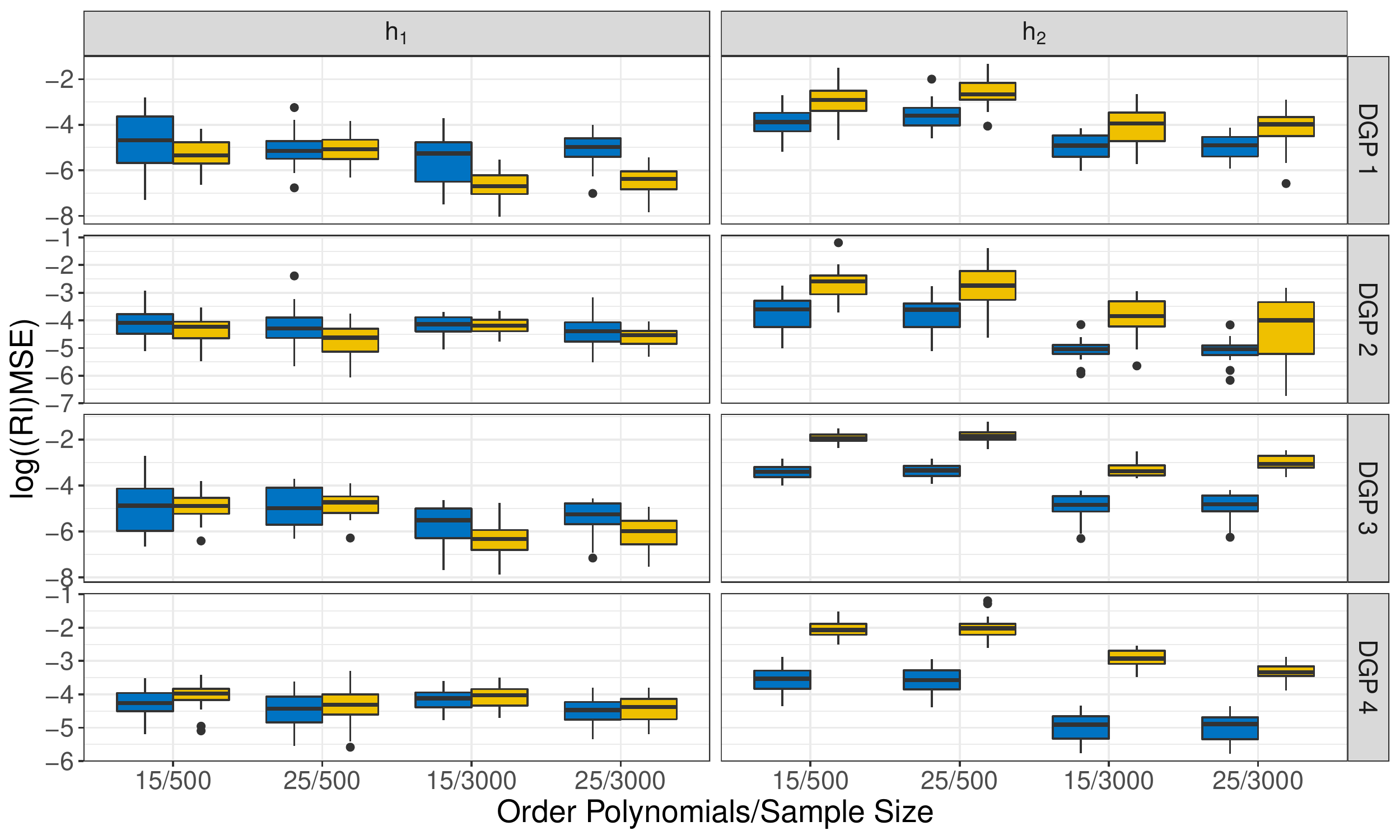}
    \caption{Comparison of the logarithmic (RI)MSEs between TBM-Shift (yellow) and DCTM (blue) for different data generating processes (DGP in rows) as well as different orders of the Bernstein polynomial and the sample size ($M$/$n$ on the x-axis) for 20 runs. The specification of the DGPs can be found in the Appendix and for this figure are based on $\sigma_1$ with alternating $g \in \{g_1,g_2\}$ and $\eta \in \{\eta_1,\eta_2\}$. %Deviations from $\beta_1$ in $\eta_2$ were assessed by the MSE while otherwise the RIMSE was used.
    }
    \label{fig:struct}
    \vskip -0.1in
\end{figure}
The simulation results for the 6 remaining DGPs can be found in the supplementary material. For $h_1$ and $h_2$, the results for the majority of specifications reveal that DCTMs benefit from lower order Bernstein polynomials independent of the sample size. When only unstructured model components were specified, DCTM's estimation of $h_1$ benefits from Bernstein polynomial with higher order. This holds regardless of $g$. %Due to the limitations of TBMs mentioned in Section \ref{sec:contribution} we were not able to specify TBMs for these 10 specification of the numerical experiments so that correctly specified models that were computationally feasible evolved.
Figure~\ref{fig:struct2} exemplary depicts the estimation performance of DCTMs for one DGP setting.
\begin{figure}[ht]
    \centering
    \includegraphics[width = \columnwidth]{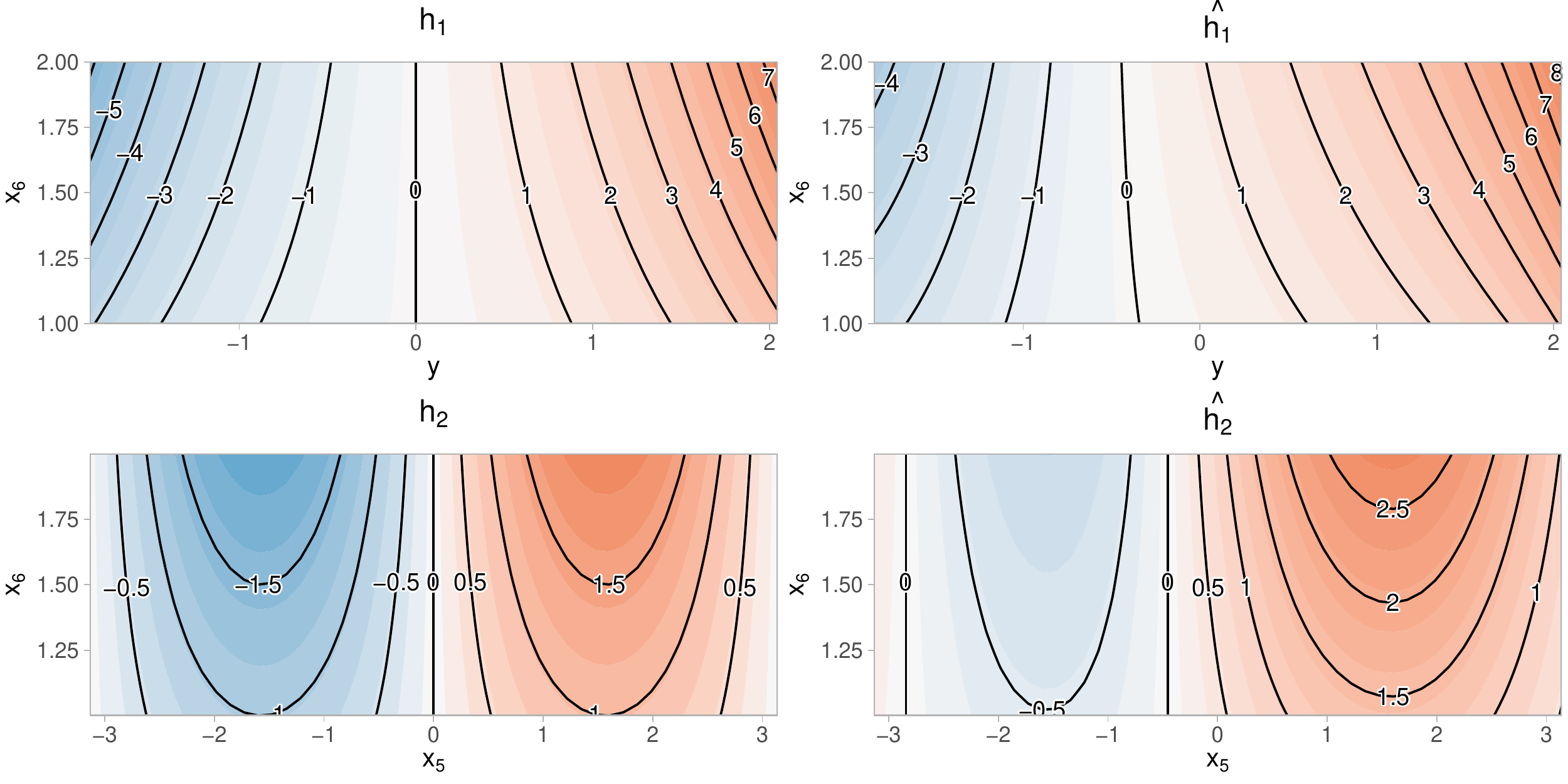}
    \caption{Exemplary visualization of the learned feature-driven interaction term $h_1$ (upper row) as well as shift term $h_2$ (lower row). Plots on the left show the data generating surface $h_1, h_2$, plots on the right the  estimated surface $\hat{h}_1, \hat{h}_2$ for different values of the feature inputs. The plots correspond to the DGP setting $g_1,\eta_1,\sigma_2$ (see Appendix).}
    \label{fig:struct2}
\end{figure}

\section{Application}

We now demonstrate the application of DCTMs by applying the approach to a movie reviews and a face data set.

\subsection{Movie Reviews}

The Kaggle movies data set consists of $n=4442$ observations. Our goal is to predict the movies' revenue based on their production budget, popularity, release date, runtime and genre(s). Figure~1 in the Appendix depicts the revenue for different genres. We deliberately do not log-transform the response, but let the transformation network convert a standard normal distribution (our error distribution) to fit to the given data.

\paragraph{Model Description}

First, we define a DCTM solely based on a structured additive predictor (i.e. no deep neural net predictor) as a baseline model which we refer to as the ``Structured Model''. The structured additive predictor includes the binary effect for each element of a set of 20 available genres ($x_0$) as well as smooth effects (encoded as a univariate thin-plate regression splines \cite{Wood.2003}) for the popularity score ($x_1$), for the difference of release date and a chosen date in the future in days ($x_2$), for the production budget in US dollars ($x_3$) and the run time in days ($x_4$): 
\begin{equation}\label{eq:linpredapp}
\begin{split}
\textstyle \sum_{r=1}^{20} &\beta_{r} I(r \in x_{0,i}) + s_{1}(x_{1,i}) + s_{2}(x_{2,i}) + s_{3}(x_{3,i}) + s_{4}(x_{4,i}).
\end{split}    
\end{equation}
This linear predictor \eqref{eq:linpredapp} is used to define the structured component in in the shift term $\beta(\bm{x})$. For the interaction term, the STAP consists of all the genre effects and the resulting design matrix $\bm{\mathcal{B}}$ is then combined with the basis of a Bernstein polynomial $\bm{\mathcal{A}}$ of order $M=25$. We compare this model with three deep conditional transformation models that use additional textual information of each movie by defining a ``Deep Shift Model'', a ``Deep Interaction Model'' and a ``Deep Combination Model''. The three models all include a deep neural network as input either in the shift term, in the interaction term or as input for both model parts, respectively. As deep neural network we use an embedding layer of dimension $300$ for $10000$ unique words and combine the learned outputs by flatting the representations and adding a fully-connected layer with 1 unit for the shift term and/or 1 units for the interaction term on top. As base distribution we use a logistic distribution, i.e., $F_Z(h) = F_L(h) = (1+\exp(-h))^{-1}$.

\paragraph{Comparisons}

We use 20\% of the training data as validation for early stopping and define the degrees-of-freedom for all non-linear structured effects using the strategy described in Section~\ref{sec:penalty}. We compare our approach again with the shift and distributional TBM (TBM-Shift and TBM-Distribution, respectively) as state-of-the-art baseline. %TBM can be used to optimize the likelihood of a distributional CTM to get an estimate for $h_1$ (TBM-Distribution) or to the likelihood of a shift CTM to get an estimate of $h_2$ (TBM-Shift). 
%The TBM framework uses a component-wise gradient boosting algorithm where each additive effect can be defined by potentially different base learners. 
We run both models with the predictor specification given in \eqref{eq:linpredapp}. %The genres were specified by linear base learners while all other covariates were modeled using univariate P-spline base learners. The main tuning parameter of the algorithm is the number of boosting iterations. 
For TBM, we employ a 10-fold bootstrap to find the optimal stopping iteration by choosing the minimum out-of-sample risks averaged over all folds. Finally we evaluate the performance on the test data for both algorithms.

\paragraph{Results}

The non-linear estimations of all models show a similar trend for the four structured predictors. Figure~\ref{fig:struct} depicts an example for the estimated partial effects in the $h_2$ term of each model. %Note, however, that the features are plotted on a log-scale and that deviations occur in particular for larger values, for which there is little data for all four features.  %Figure~\ref{fig:deep} depicts the estimated densities for different observations in the test set based on the deep shift model.
The resulting effects in Figure~\ref{fig:struct} can be interpreted as functional log-odds ratios due to the choice $F_Z = F_L$. For example, the log-odds for higher revenue %and thus the effect on the entire $\hat{F}_{Y|\bm{x}}$ 
linearly increase before the effect stagnates for three of the four model at a level greater than 150 million USD.
\begin{figure}
\begin{center}
\hspace*{-0.7cm}    \includegraphics[width = 0.6\columnwidth]{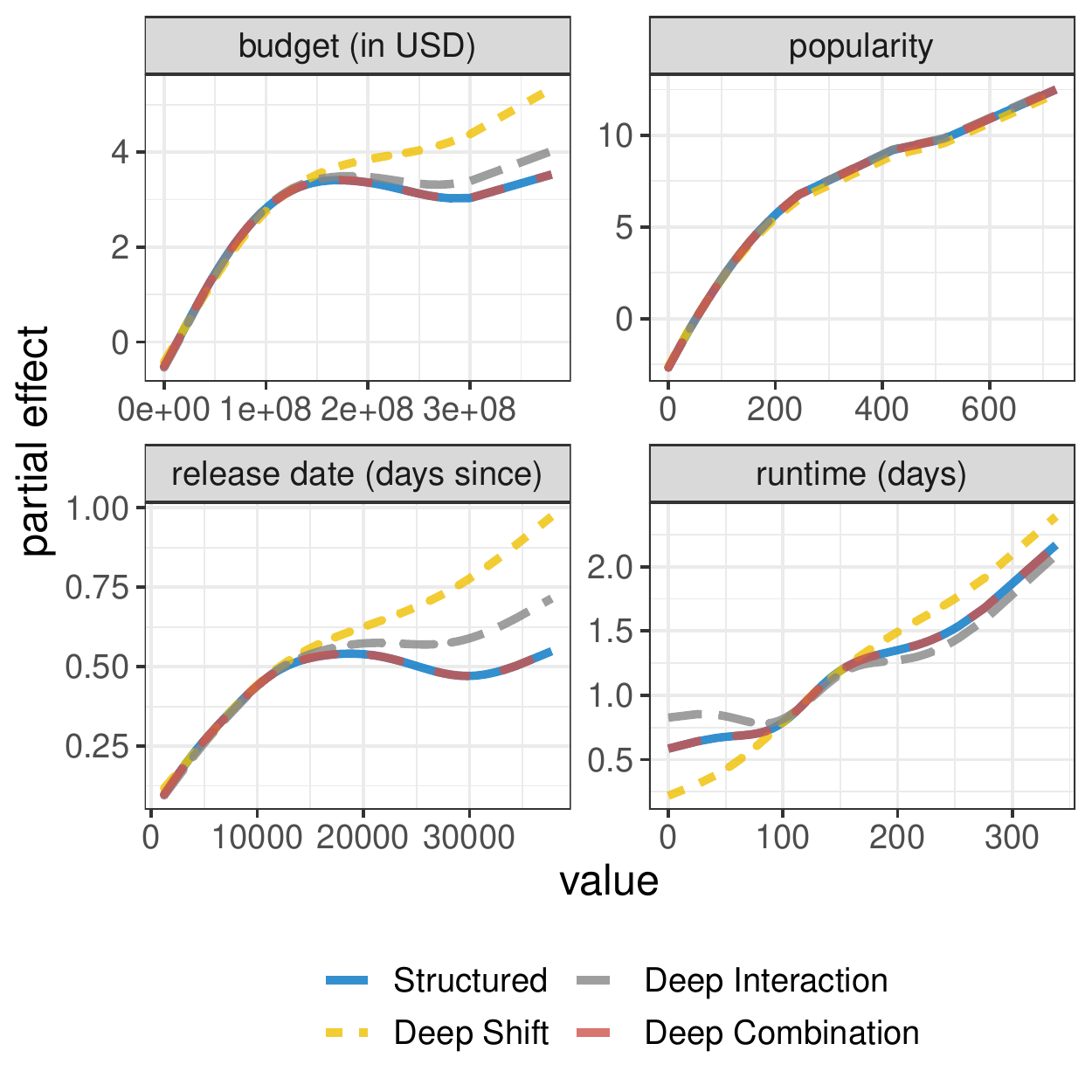}
\vspace*{-0.1cm}
    \caption{Estimated non-linear partial effect of the 4 available numerical features for $h_2$ (in each sub-plot) based on the four different DCTM models (colors).}
    \label{fig:struct}
    \end{center}
    \vskip -0.2in
\end{figure}
Table~\ref{tab:deep} shows (Movie Reviews column) the mean predicted log-scores \cite{Gelfand.1994}, i.e., the average log-likelihood of the estimated distribution of each model when trained on 80\% of the data (with 20\% of the training data used as validation data) and evaluated on the remaining 20\% test data. Results suggest that deep extensions with movie descriptions as additional predictor added to the baseline model can improve over the TBM, but do not achieve as good prediction results as the purely structured DCTM model in this case. Given the small amount of data, this result is not surprising and showcases a scenario, where the potential of the structured model part outweighs the information of a non-tabular data source. The flexibility of our approach in this case allows to seamlessly traverse different model complexities and offers a trade-off between complexity and interpretability.
\renewcommand{\arraystretch}{1.1}
\begin{table}[htbp]
\vskip -0.1in
\begin{small}
\begin{center}
\label{tab:deep}
\caption{Average result (standard deviation in brackets) over different training/test-splits on the movie reviews (left) and UTKFace data set. Values correspond to negative predicted log-scores (PLS; smaller is better) for each model with best score in bold.}
\vspace{3pt}
\begin{tabular}{p{8pt}lcc}
& Model & Movie Reviews & UTKFace\\
\hline
\multirow{4}{*}{\rotatebox[origin=c]{90}{DCTM}} & Structured & \textbf{19.26} (0.18) & 3.98 (0.02)\\
& Deep Shift & 19.32 (0.20) & 3.81 (0.52) \\
& Deep Interaction & 19.69 (0.22) & 3.79 (0.21)\\
& Deep Combination & 19.67 (0.19) & \textbf{3.37} (0.09)\\ \hline
\multirow{2}{*}{\rotatebox[origin=c]{90}{TBM}} & Shift & 23.31 (0.83) & 4.25 (0.02)\\
& Distribution & 22.38 (0.31) & 4.28 (0.03) \\
\hline
\end{tabular}
\end{center}
\end{small}
\vskip -0.2in
\end{table}

\subsection{UTKFace}

The UTKFace dataset is a publicly available image dataset with $n=23708$ images and additional tabular features (age, gender, ethnicity and collection date). We use this data set to investigate DCTMs in a multimodal data setting. 

\paragraph{Model Description}

Our goal is to learn the age of people depicted in the images using both, the cropped images and the four tabular features. As in the previous section we fit the four different DCTM models, all with the same structured additive predictor (here effect for race, gender and a smooth effect for the collection date) and add a deep neural network predictor to the $h_1$ (Deep Interaction), $h_2$ (Deep Shift), to both (Deep Combination) or only fit the structured model without any information of the faces (Structured). The architecture for the faces consists of three CNN blocks (see Appendix for details) followed by flattening operation, a fully-connected layer with 128 units with ReLU activation, batch normalization and a dropout rate of 0.5. Depending on the model, the final layer either consists of 1 hidden unit (Deep Shift, Deep Interaction) or 2 hidden units (Deep Combination).

\paragraph{Comparisons}

The baseline model is a two-stage approach that first extracts latent features from the images using a pre-trained VGG-16 \cite{Zisserman.2015} and then uses these features together with the original tabular features in a TBM-Shift/-Distribution model to fit a classical structured additive transformation model. We again compare the 4 DCTM models and 2 baseline models using the PLS on 30\% test data and report model uncertainties by repeating the data splitting and model fitting 4 times. For the DCTMs we use early stopping based on 20\% of the train set used for validation. For TBM models we search for the best stopping iteration using a 3-fold cross-validation. The results in Table~\ref{tab:deep} (UTKFace column) suggest that our end-to-end approach works better than the baseline approach and that the DCTM benefits from a combined learning of $h_1$ and $h_2$ through the images.

\subsection{Benchmark Study}

We finally investigate the performance of our approach by comparing its density estimation on four UCI benchmark data sets (Airfoil, Boston, Diabetes, Forest Fire) against parametric alternatives. We use a deep distributional regression approach (DR) \cite{Ruegamer.2020}, a Gaussian process (GP) and a GP calibrated with an isotonic regression~ (IR) \cite{kuleshov18a}. We adapt the same architecture as in DR to specifically examine the effect of the proposed transformation. To further investigate the impact  of the polynomials' order M (i.e., flexibility of the transformation vs. risk of overfitting), we run the DCTM model with $M\in\{1,16,32,64\}$ (DCTM-$M$). We also include a normalizing flow baseline with a varying number of radial flows $\mathcal{M}$ (NF-$\mathcal{M}$; \cite{Rothfuss.2020}). This serves as a reference for a model with more than one transformation and thus potentially more expressiveness at the expense of the feature-outcome relationship being not interpretable. Details for hyperparameter specification can be found in the Appendix.
\renewcommand{\arraystretch}{1}
\setlength{\tabcolsep}{4pt}
\begin{table}[htbp]
\vskip -0.1in
\caption{Comparison of neg. PLS (with standard deviation in brackets)
%(first four rows) and MSE (second four rows) 
of different methods (rows; best-performing model in bold, second best underlined) on four different UCI repository datasets (columns) based on 20 different initializations of the algorithms.}
\label{uci}
\begin{center}
\begin{footnotesize}
\begin{tabular}{ccccc}
     & Airfoil & Boston & Diabetes & Forest F. \\ \hline
DR & {3.11} (0.02) & 3.07 (0.11) & \textbf{5.33} (0.00) & 1.75 (0.01) \\
GP   & 3.17 (6.82) & 2.79 (2.05) & 5.35 (5.76) & 1.75 (7.09) \\
IR   & 3.29 (1.86) & 3.36 (5.19) & 5.71 (2.97) & \textbf{1.00} (1.94) \\ \hline
DCTM-1   & 3.07 (0.01) & 2.97 (0.03) & 5.44 (0.02) & 1.83 (0.02) \\
% DCTM-4   & 3.07 (0.02) & 2.85 (0.03) & 5.35 (0.01) & 1.77 (0.06) \\
%DCTM-6   & &  &  & \\
%DCTM-8  & 3.07 (0.02) & 2.79 (0.02) &  5.37 (0.01) & 1.56 (0.08) \\
DCTM-16  & 3.07 (0.02) & 2.76 (0.02) & \underline{5.34} (0.01) & 1.30 (0.12) \\
DCTM-32  & 3.08 (0.02) &  \underline{2.71} (0.03) & 5.39 (0.02) & \underline{1.08} (0.15)  \\
DCTM-64  & 3.08 (0.03) & \textbf{2.66} (0.05) & 5.37 (0.01) & 1.41 (1.03) \\ \hline
NF-1 & 3.04 (0.22) & 2.98 (0.20) & 5.59 (0.10) & 1.77 (0.02) \\
%NF-2 & 2.95 (0.21) & 2.81 (0.17) & 5.57 (0.12) & 1.77 (0.02) \\
NF-3 & \textbf{2.88} (0.17) &  2.76 (0.14) & 5.54 (0.11) & 1.76 (0.02)\\
%NF-4 & 2.87 (0.21) & 2.75 (0.13) & 5.54 (0.12) & 1.75 (0.01)  \\
NF-5 & \underline{2.90} (0.18) & 2.81 (0.20) & 5.47 (0.09) & 1.77 (0.12)
\end{tabular}
\end{footnotesize}
\end{center}
\vskip -0.2in
\end{table}
Results (Table~\ref{uci}) indicate that our approach performs similar to alternative methods. For two data sets, the greater flexibility of the transformation yields superior performance compared to methods without transformation (DR, GP, IR), suggesting that the transition from a pure parametric approach to a more flexible transformation model can be beneficial. For the other two data sets, DCTM's performance is one standard deviation apart from the best performing model. For Airfoil the even greater flexibility of a chain of transformations (NF-$\mathcal{M}$ in comparison to DCTM-$M$) improves upon the result of DCTMs.

\section{Conclusion and Outlook}
We introduced the class of deep conditional transformation models which unifies existing fitting approaches for transformation models with both interpretable (non-)linear model terms and more complex predictors in one holistic neural network. A novel network architecture together with suitable constraints and network regularization terms is introduced to implement our model class. Numerical experiments and applications demonstrate the efficacy and competitiveness of our approach. 

%For future research we will investigate the extension of our approach. One possibility is to extend the framework to multivariate outcomes. Our framework can also be easily extended to discrete ordinal or interval censored outcomes, as the conditional transformation model specified in \eqref{eq:CTM} and the basis function specification for the transformation function $h(y) = \bm{a}(y)^\top \bm{\vartheta}$ can also be defined for discrete $y$. 

\subsubsection*{Acknowledgements}
This work has been partly funded by SNF grant 200021-184603 from the Swiss National Science Foundation (Torsten Hothorn) and the German Federal Ministry of Education and Research (BMBF) under Grant No. 01IS18036A (David R\"ugamer).

\bibliographystyle{splncs04}
\bibliography{bibliography}
%Author, F.: Article title. Journal \textbf{2}(5), 99--110 (2016)

%Author, F., Author, S.: Title of a proceedings paper. In: Editor,
%F., Editor, S. (eds.) CONFERENCE 2016, LNCS, vol. 9999, pp. 1--13.
%Springer, Heidelberg (2016). \doi{10.10007/1234567890}

%Author, F., Author, S., Author, T.: Book title. 2nd edn. Publisher,

%Author, A.-B.: Contribution title. In: 9th International Proceedings
%on Proceedings, pp. 1--2. Publisher, Location (2010)

%LNCS Homepage, \url{http://www.springer.com/lncs}. Last accessed 4
%\end{thebibliography}
\end{document}